\documentclass[conference]{IEEEtran}
\IEEEoverridecommandlockouts
\usepackage{cite}
\usepackage{amsmath,amssymb,amsfonts}
\usepackage{algorithmic}
\usepackage{graphicx}
\usepackage{textcomp}
\usepackage{xcolor}
\def\BibTeX{{\rm B\kern-.05em{\sc i\kern-.025em b}\kern-.08em
    T\kern-.1667em\lower.7ex\hbox{E}\kern-.125emX}}
    
\usepackage{amsmath}
\usepackage{algorithm}
\usepackage{algorithmic}
\usepackage{hyperref}
\usepackage{tikz}
\usepackage{float}
\def\checkmark{\tikz\fill[scale=0.4](0,.35) -- (.25,0) -- (1,.7) -- (.25,.15) -- cycle;} 
\usepackage{graphicx}
\usepackage{xspace}

\usepackage{multirow}
\begin{document}

\title{

Electrical Load Forecasting in Smart Grid: \\ A Personalized Federated Learning Approach
}

\author{
        \IEEEauthorblockN{
        Ratun Rahman\IEEEauthorrefmark{2}, Neeraj Kumar\IEEEauthorrefmark{3}, and Dinh C. Nguyen\IEEEauthorrefmark{2}
	}

	\IEEEauthorblockA{
	\IEEEauthorrefmark{2}Department of Electrical and Computer Engineering, University of Alabama in Huntsville, AL, USA \\
 \IEEEauthorrefmark{3}Pacific Northwest National Laboratory, Richland, WA, USA
	}
 
 Emails: rr0110@uah.edu, neeraj.kumar@pnnl.gov, dinh.nguyen@uah.edu}
	\markboth{}%
	{}

\maketitle
\pagenumbering{gobble} 



\begin{abstract}
Electric load forecasting is essential for power management and stability in smart grids. This is mainly achieved via advanced metering infrastructure, where smart meters (SMs) are used to record household energy consumption. Traditional machine learning (ML) methods are often employed for load forecasting but require data sharing which raises data privacy concerns. Federated learning (FL) can address this issue by running distributed ML models at local SMs without data exchange. However, current FL-based approaches struggle to achieve efficient load forecasting due to imbalanced data distribution across heterogeneous SMs. This paper presents a novel personalized federated learning (PFL) method to load prediction under non-independent and identically distributed (non-IID) metering data settings. Specifically, we introduce meta-learning, where the learning rates are manipulated using the meta-learning idea to maximize the gradient for each client in each global round. Clients with varying processing capacities, data sizes, and batch sizes can participate in global model aggregation and improve their local load forecasting via personalized learning. Simulation results show that our approach outperforms state-of-the-art  ML and FL methods in terms of better load forecasting accuracy.

\end{abstract}

\maketitle

\begin{IEEEkeywords}
Federated learning, personalized federated learning, load forecasting, smart meter
\end{IEEEkeywords}

\section{Introduction}

Electrical load forecasting is crucial for power management in smart grids. This service is mainly supported via advanced metering infrastructure, where smart meters (SMs) are used to record household energy consumption and share this data with the utility \cite{9770488}. This enables utility providers to estimate future electricity demands and thereby bolster grid reliability. Conventional load-forecasting techniques in machine learning (ML) and deep learning (DL) techniques utilize pattern-finding abilities to predict future outcomes. For example, long short-term memory (LSTM) has shown its potential for time-series data-based load forecasting applications \cite{hong2020deep, bouktif2018optimal}. Generally, these methods require that every SM sends energy-usage information to the utility company. However, this data sharing may reveal customers' sensitive information such as energy usage routines. In 2009, the compulsory roll-out of SMs in the Netherlands was halted following a court ruling that the metering data collection violated customers’ privacy rights \cite{cuijpers2013smart}.

Federated learning (FL) addresses data-sharing problems by introducing a collaborative model training solution over clients and a single server \cite{nguyen2021federated1}. Here local model is created using local SM's data and sent to the global server for the next global round \cite{fekri2022distributed, taik2020electrical}. This also enables complex and large data handling which leads to quicker and more accurate communication. However, it struggles with heterogeneous data and assumes every client has the same data and capabilities. However, each client is represented as an SM in load prediction, and since they are typically highly distinct from one another, FL often ends up with an over-fitting model. 

In order to address this problem, we provide a novel method for load prediction in this research by implementing personalized federated learning (PFL). PFL removes data over-fitting by creating a custom model for every client. Our PFL technique is meta-learning which helps local models to be trained properly by choosing the best parameters using the trial and run method \cite{9428530}. As a result, it allows clients who may have poor performance due to their data quality to be trained and adjusted. The major advantages of meta-learning over other PFL techniques are that it does not add any additional calculations and costs to the server and can easily scale, making the global model aggregation fast and reliable. 

\textbf{Related Work:} There were various techniques used for load prediction. The works in \cite{rafi2021short, bouktif2018optimal} proposed an LSTM-based model to estimate electrical load demands in smart grids; \textit{however, this technique requires data sharing and hence sensitive user information such as energy consumption patterns may be exposed to third parties.} The authors in \cite{taik2020electrical, briggs2022federated} demonstrated that using FL could further improve accuracy without compromising data privacy. Furthermore, it was also able to reduce significant networking load. Another study in \cite{fekri2022distributed} also showed similar performance. They further added that FedAVG, which carried out several steps before merging updates on the server performed better than other FL techniques like FedSGD. \textit{However, they struggle with non-IID data where data distributions are imbalanced across heterogeneous SMs. \cite{rahman2024improved}}

Recently, PFL techniques have been considered to tackle the data heterogeneity issue in load forecasting. The study in \cite{9770488} proposed a PFL technique for load forecasting where each SM customizes a federated prediction model using its own data. Another work in \cite{10233242} introduced a Generative Adversarial Network (GAN) based differential privacy (DP) algorithm that included multi-task PFL. \textit{However, this solution increases computational complexity at the server and reduces the scalability of the load-forecasting network.} We compare our approach with related works in Table~\ref{table:related_work_table}. 


\begin{table}
\footnotesize
\centering
\caption{Comparison of our approach with existing methods.}
\begin{tabular}{|p{2cm}||p{0.6cm}|p{0.6cm}|p{0.6cm}|p{0.6cm}|p{0.6cm}|p{0.6 cm}|p{09.8cm}|}
 \hline
 Objectives& \cite{rafi2021short, bouktif2018optimal} &  \cite{taik2020electrical, fekri2022distributed, briggs2022federated} & \cite{9770488} & \cite{10233242} & Our Approach\\
 \hline
 Handle uncertain and non-iid data & & & \checkmark& \checkmark & \checkmark\\
 \hline
 Include diverse SMs& \checkmark & & \checkmark &  \checkmark & \checkmark\\
 \hline
 Adaptability to user change& \checkmark & \checkmark &  & \checkmark & \checkmark\\
 \hline
 Handle large dataset & & \checkmark & \checkmark & \checkmark& \checkmark\\
 \hline
 Maintain server complexity & & \checkmark & \checkmark & &\checkmark\\
 \hline
 Keeping data secured & & \checkmark & \checkmark & \checkmark& \checkmark\\
 \hline
  \textcolor{black}{Latency minimization} & &  &  & & \checkmark\\
 \hline
\end{tabular}
\label{table:related_work_table}
\vspace{-5mm}
\end{table}

\textbf{Our Key Contributions:} Motivated by the above limitations, \textit{we propose a new PFL method for high-quality load forecasting over SMs in the smart grid}. Specifically, inspired by gradient compression methods in \cite{karimireddy2019error, vogels2019powersgd}, we incorporate different gradients to build a more personalized local load forecasting model at each SM. This is more widely known as \textit{meta learning} which introduces the 'learning to learn' concept for PFL \cite{fallah2020personalized}. To this end, our key contributions are summarized as follows:
\begin{itemize}
\item We propose a new PFL approach called personalized meta-LSTM algorithm with a flexible SM participation method for collaborative load forecasting in the smart grid. This allows complicated and diverse data to be structured, assembled, and processed quickly and removes data sharing to protect the privacy and security of household electricity recordings.  
\item For each client, we put adaptive model aggregation. Prior to training the local model, the clients are temporarily evaluated using varying learning rates. The most suitable learning rate is then selected among the available learning rates based on which one yields the lowest loss value. Next, we move forward with training local models with the optimal learning rate.
\item We carry out extensive simulations on real-world datasets and compare our model's performance for both IID and non-IID settings with existing methods. 
\end{itemize}

The rest of the paper is as follows. We present our system model in Section II. In Section III, we discuss the problem formulation. We evaluate our simulation results and performance evaluation in Section IV. Section V concludes the paper. 

\section{System Model}

\subsection{Overall System Architecture}
\begin{figure}[!t]
\centering
\includegraphics[width=3.2in]{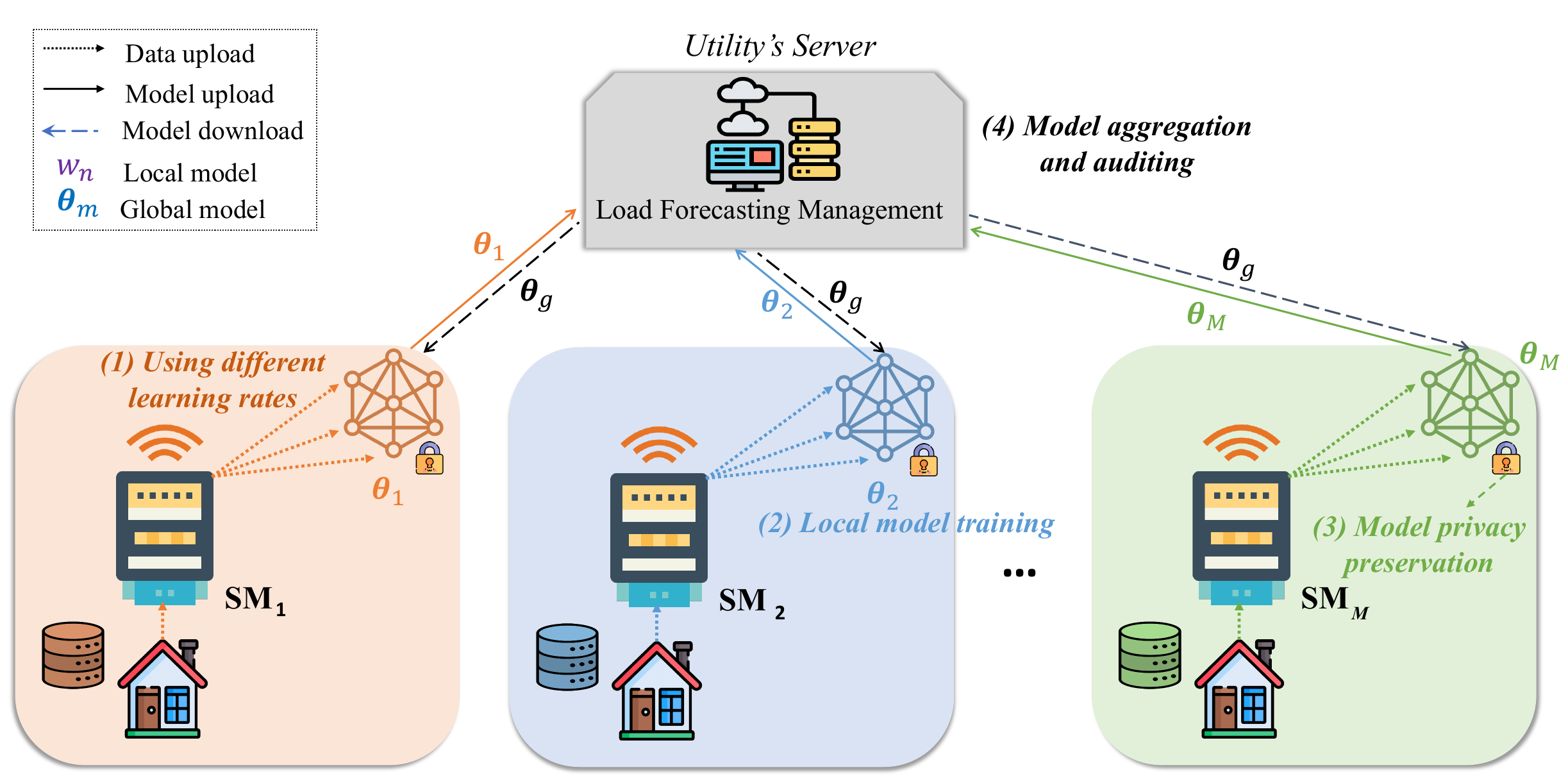}
\caption{Our proposed architecture in load forecasting network.}
\vspace{-1mm}
\label{Fig: Overview}
\vspace{-5mm}
\end{figure}

Fig.~\ref{Fig: Overview} illustrates the overall system and architecture for our proposed architecture. Here, a utility server (US) is considered a global server where global model aggregation is performed based on a shared local model in every global round for load prediction. Each SM (client), denoted as $m \in \mathcal{M}$,  records household energy consumption data. Energy recordings are time-varying, and differ over SMs. We denote every global round as $k \in \mathcal{K}$ where $k=\{1,2,3,\dots, K\}$ and $K$ is the final global round. In each round $k$, each SM $m$ contains a local dataset $D_m^{(k)}$ that varies in every round and client, with size $D_m^{(k)} = |D_m^{(k)}|$. The SMs employ these datasets for global training in round $k$ and the total dataset is $D^{(k)} = \sum_{m\in \mathcal{M}} D_m^{(k)}$. 

The goal of the system is to train each local model $\theta_m^{(k)}$ effectively such that the global model $\theta_g^{(k+1)}$, created by their aggregation, can provide better results. To create a local model, each SM requires a gradient parameter denoted by $\nabla F$ with a learning rate $(\alpha)$ to fasten or slow the learning process. For our personalized approach, we have a series of learning rates ($\alpha_1,\alpha_2,\dots$) and we calculate the loss value for every learning rate for each SM $m$ in each round $k$ in a small dataset (Dataloader). The learning rate that provides the lowest loss value is considered the optimal learning rate $(\beta_m^{(k)})$. Then we use $\beta_{m,k}$ for the local model training and send the local weight to the server for global model aggregation. Then the global model is updated for the next training round $k+1$ using the basic federated averaging.

In our approach, we have used 80\% of our data for local model training and expressed them as Trainloader $(D_{m,k}^{train})$ and the rest 20\% for testing as Testloader $(D^{test})$. To evaluate the initial performance for meta-learning, we have used 20\% of the training data expressed as Dataloader $(D_{m,k}^{temp})$ and used that to find $\beta_{m,k}$.

\begin{algorithm}
\footnotesize
	\caption{{Proposed PFL algorithm for high-quality load forecasting across SMs}}
	\begin{algorithmic}[1]
		\label{algo: metaSGD}
		\STATE \textbf{Input:}  The set of global communication rounds $\mathcal{K}$, local training round $\mathcal{T}$, a set of SMs $\mathcal{M}$
		\STATE \textbf{Initialization:} Initialize global model $\boldsymbol{w}_0$, different learning rates $\alpha_{0,1,\dots,j}$
		\FOR{each global communication round $k \in \mathcal{K}$}
		\STATE Send $\boldsymbol{w}_k$ to sampled SMs
		\FOR{each sampled SM $m \in \mathcal{M}$ in parallel}
		\FOR{each local training epoch $t \in \mathcal{T}$}
        \STATE Get $\boldsymbol{w}_k$ 
        \FOR{each learning rates $\alpha_i$ where $i \in j$}
        \STATE Calculate $f_{i,k}^{'} (\alpha_i) = f_{i,k}((\boldsymbol{w}_{m,k}^t,\alpha_i), D_{m,k}^{temp})$ on $D_{m,k}^{temp})$
        \STATE Save the best learning rate as $(\beta_{m,k})$ that has the lowest $f_i^{'}$
        \STATE Return $\beta_{m,k}$ to the local model $\theta_m$
        \ENDFOR
        \STATE Perform local model training (meta-learning) on $\theta_i$, $\boldsymbol{w}_{m,k}^{t+1} = \boldsymbol{w}_{m,k}^{t} - \beta_{m,k} \nabla F(\boldsymbol{w}_{m,k}^t,D_{m,k}^{train})$
		\ENDFOR
        \STATE Send $\boldsymbol{w}_{m,k}$ to the server
		\ENDFOR
		\STATE The utility's server updates the global parameter by averaging: $\boldsymbol{w}_{k+1} = \frac{1}{M} \sum_{m\in\mathcal{M}}\boldsymbol{w}_{m,k}$ 
        \STATE Perform test on the updated weight $\boldsymbol{w}_{k+1}$
		\STATE The utility's server  broadcasts the aggregated global model $\boldsymbol{w}_{k+1}$ to all participating SMs for the next round of training
		\ENDFOR
        \STATE \textbf{Output:} Optimal global load forecasting model $\boldsymbol{w}^*$
  \end{algorithmic}
\end{algorithm}

\subsection{Objective Function}
The goal of our proposed approach is to learn which learning rate to use for the local model training by minimizing the following objective function:
\begin{equation}
    \min Fed_{avg} = \frac{1}{\mathcal{K}} \sum_{k=1}^{\mathcal{K}} \mathcal{L}(y^k,\hat y^k),
\end{equation}
where $\mathcal{L}$ is a loss function. $y^k$ represents the actual value, and $\hat y^k$ is the predicted value for the $k^{th}$ task. 

Load forecasting is a regression-based task. As a result, we train the model to minimize the mean absolute error (MAE) given by:
\begin{equation}
    \mathcal{L}(y^k,\hat y^k) = \frac{1}{n} \sum_{i=1}^{n} |y_i - \hat{y}_i|.
\end{equation}

We have also considered the root mean squared error (RMSE) using the equation below
\begin{equation}
    \mathcal{L}(y^k,\hat y^k) = \sqrt{\frac{1}{N} \sum_{i=1}^N (y_i^k-\hat y_i^k)^2}.
\end{equation}

\subsection{LSTM}
\begin{figure}[!t]
\centering
\includegraphics[width=3.4in]{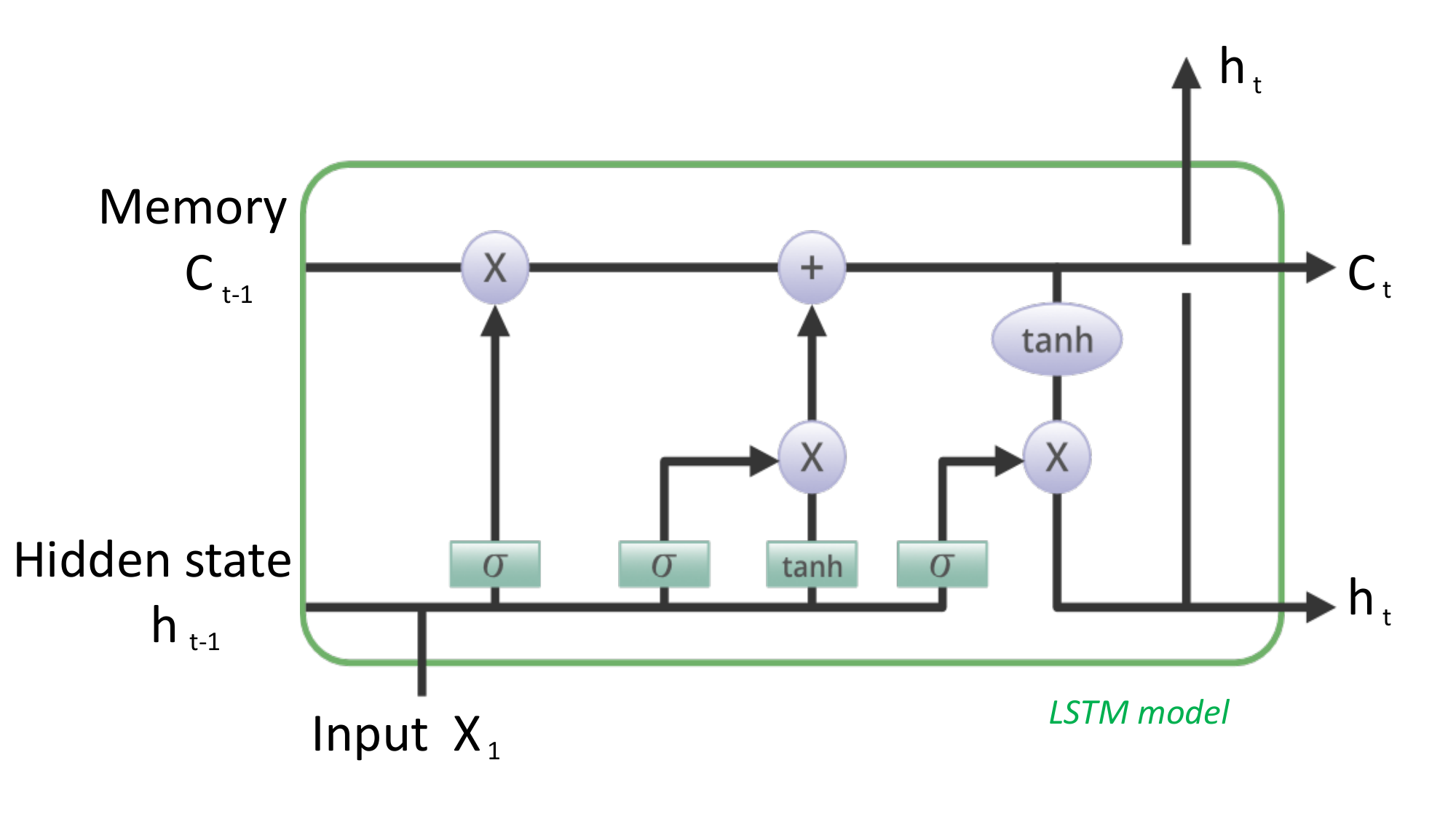}
\vspace{-5mm}
\caption{LSTM model architechture.}
\label{Fig: lstm}
\vspace{-5mm}
\end{figure}

LSTM is usually used in time sequences and long-range dependencies data. To forecast future values based on past data, load prediction usually involves finding patterns and trends across time. As a result, LSTM is well-suited for load prediction \cite{kong2017short}. LSTM is a form of recurrent neural network (RNN) architecture that consists of unique units or memory cells that are designed to retain their state over time and regulate the processing, processing, and storage of information. Each LSTM unit has three different sorts of gates that facilitate this: input, forget, and output gates. It also has two other components: cell state $c_t$ represents internal memory and hidden state $h_t$ represents the output of the LSTM unit at time step t. Fig. \ref{Fig: lstm} describes the basic LSTM architecture. 
At each time step $t$, LSTM does the following operation:
\begin{enumerate}
    \item What data to remove from the cell state is determined by the forget gate ($f_t$).
    \begin{equation}
        f_t = \sigma(W_f.[h_{t-1},x_t]+b_f)
    \end{equation}
    where $\sigma$ is the sigmoid activation function, $W_f$ and $b_f$ are the weight matrix and bias for the forget gate $f_t$, $h_{t-1}$ is the hidden state from the previous time step, and $x_t$ is the input at the current time step.
    \item The input gate ($i_t$) determines what additional data should be added to the cell state.
    \begin{equation}
        i_t = \sigma(W_i.[h_{t-1},x_t]+b_i)
    \end{equation}
    \item A fresh candidate value to be added to the cell state is provided by the candidate cell state $\tilde C_t$.
    \begin{equation}
        \tilde C_t = tanh(W_c.[h_{t-1},x_t]+b_c)
    \end{equation}
    where $tanh$ is the hyperbolic tangent activation function and $W_c$ and $b_c$ are the weight matrix and bias for the candidate cell state $\tilde C_t$. 
    \item Cell state update combines the old cell state, forget gate output, input gate output, and candidate cell state to change the cell state.
    \begin{equation}
        C_t =  f_t.C_{t-1} + i_t.\tilde C_t
    \end{equation}
    where $C_{t-1}$ is the cell state from the previous time step. 
    \item From the current cell state output gate $o_t$ decides what information to output.
    \begin{equation}
        o_t = \sigma(W_o.[h_{t-1},x_t]+b_o)
    \end{equation}
    \item The hidden state $h_t$ is updated for the current time step.
    \begin{equation}
        h_t = o_t.tanh(C_t)
    \end{equation}
\end{enumerate}
These steps occur on every time step $(t)$. 

\section{Problem Formulation}
We assume that there are $N$ clients and a global model is $\boldsymbol{w}$. A generalized FL is given as
\begin{equation}
\min_{\boldsymbol{w}_k \in {\rm I\!R}^d} f(\boldsymbol{w}_k) := \frac{1}{N} \sum_{i=1}^N f_i(\boldsymbol{w}_k),
\end{equation}
which is used to find $\boldsymbol{w}_k$. Here the function $f_i: \rm I\!R^d \in \rm I\!R, i=1,2,3,\dots,N$ denotes the predicted loss value over $i^{th}$ client's data distribution:
\begin{equation}\label{supervised_ML}
f_i(\boldsymbol{w}_k) := {\rm I\!E}_{\xi_i} \left [ {f_i}^{'} (\boldsymbol{w}_k, x_i) \right ].
\end{equation}
In this equation ${f_i}^{'} (\boldsymbol{w}_k,x_i)$ is a loss function calculating the difference between data sample $x_i$ and its corresponding $\boldsymbol{w}_k$.

In our approach, we try to regularize the loss function based on the performance of clients. So, our approach is:
\begin{equation}
\min_{\boldsymbol{w}_k \in {\rm I\!R}^d} f(\boldsymbol{w}_k) := \frac{1}{N} \sum_{i=1}^N F_i(\boldsymbol{w}_k).
\end{equation}
where $F_i(\boldsymbol{w}_k) = \min_{\boldsymbol{w}_k \in {\rm I\!R}^d} \left\{ f_i(\boldsymbol{w}_k, x_i) \right\}$.

Here, $F_i(\boldsymbol{w}_k)$ is well known as Moreau envelope, where the optimal personalized model is defined \cite{lin2018catalyst} as follows:
\begin{equation}
    \hat{\boldsymbol{w}}_{i,k} = \text{prox}_{f_i / x_i}(\boldsymbol{w}_k) 
\end{equation}
\begin{equation}
    \hat{\boldsymbol{w}}_{i,k} = \arg \min_{\theta \in  {\rm I\!R}^d} \left\{ f_i(\boldsymbol{w}_k) + x_i \right\}.
\end{equation}

The closest formulation to our approach is pFedMe \cite{t2020personalized} where they defined the equation as:
\begin{equation}
    \boldsymbol{w}_{i,k} = \boldsymbol{w}_i - \alpha \nabla f_i(\boldsymbol{w}_k) 
\end{equation}
\begin{equation}
    \boldsymbol{w}_{i,k} = \arg\min_{\theta \in {\rm I\!R}^d} \left\{ (\nabla f_i(\boldsymbol{w}_k), \theta_i - \boldsymbol{w}_k ) + \frac{1}{2\alpha} \| \theta - \boldsymbol{w}_k \|^2 \right\},
\end{equation}

Here, $\nabla f(\boldsymbol{w}_k)$ represents the gradient of the function $f$ at point $\boldsymbol{w}_k$ and $\alpha$ is the learning rate. So the objective is to minimize the squared norm of the gradient of a function $f(\boldsymbol{w}_k)$ where $\boldsymbol{w}_k=(\boldsymbol{w}_{1,k},  \boldsymbol{w}_{2,k}, \boldsymbol{w}_{3,k}, \dots, \boldsymbol{w}_{n,k})$.

\begin{figure*}[!t]
  \centering
  \begin{tabular}{ c @{\hspace{3pt}} c }
    \includegraphics[width=.99\columnwidth]{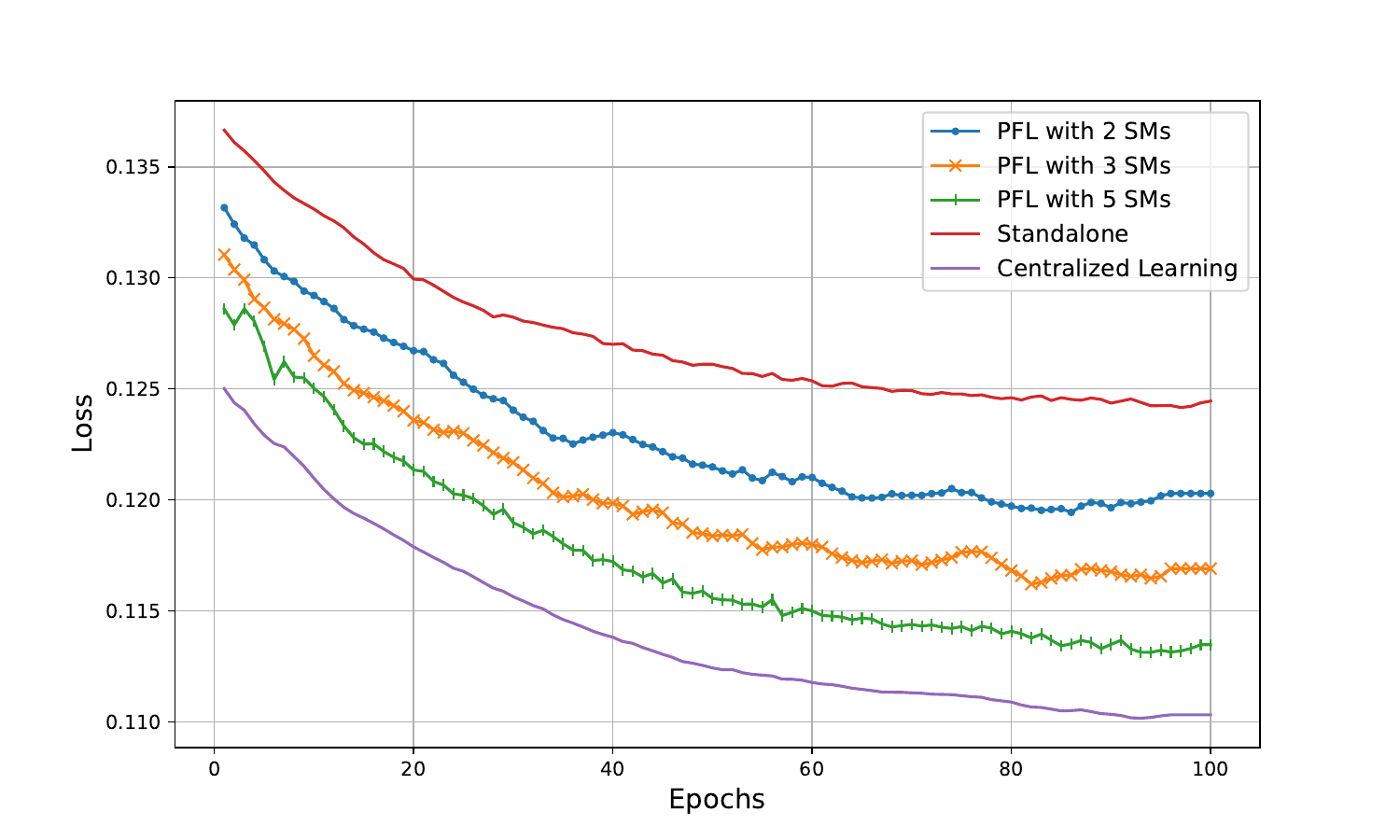} &
      \includegraphics[width=.99\columnwidth]{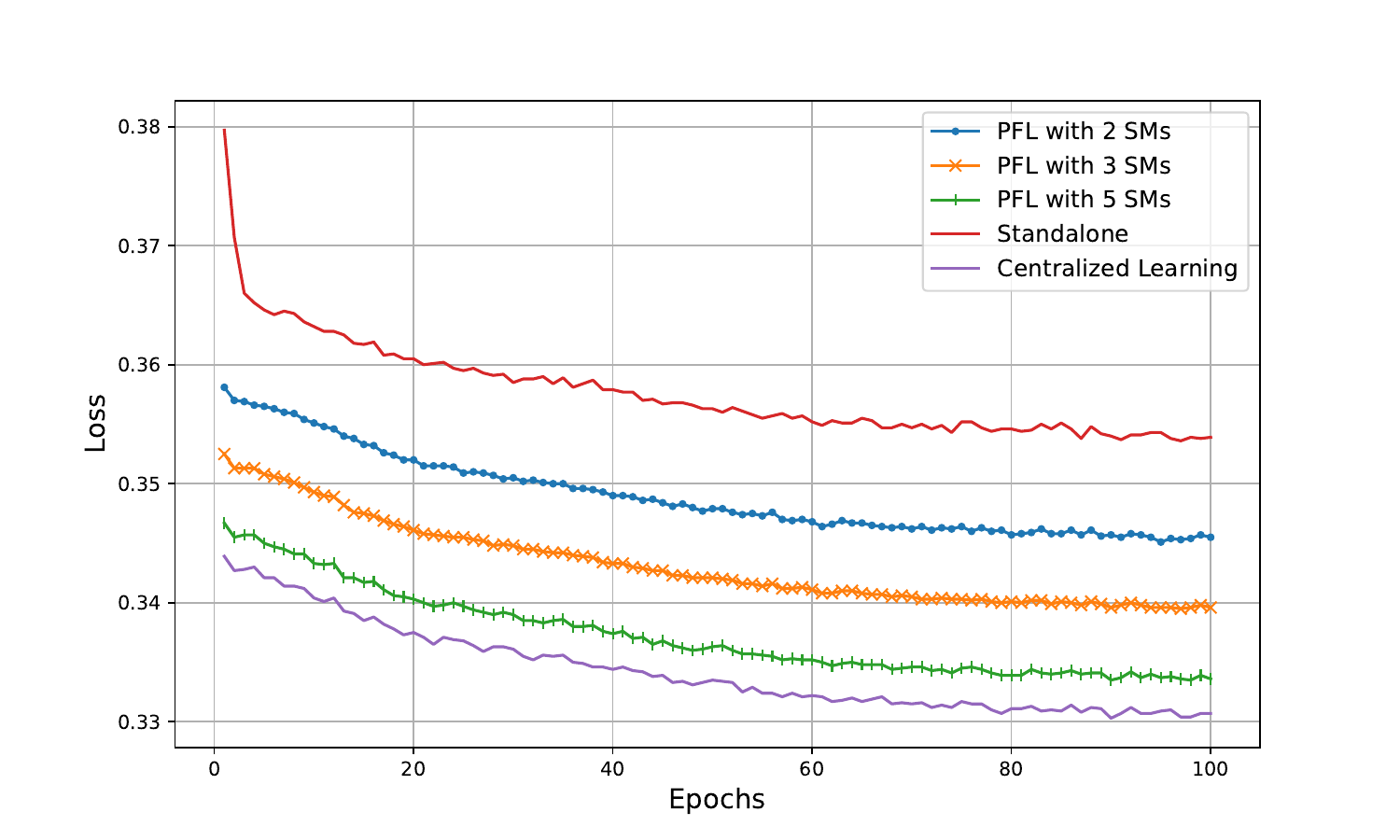} \\
    \small (a) MAE Loss &
      \small (b) RMSE Loss\\
  \end{tabular}
  \medskip
  \vspace{-5mm}
  \caption{Comparison between different numbers of clients, standalone, and centralized scheme for IID data.}
  \label{fig: iid}
  \vspace{-5mm}
\end{figure*}
\begin{figure*}[htb]
  \centering
  \begin{tabular}{ c @{\hspace{3pt}} c }
    \includegraphics[width=.99\columnwidth]{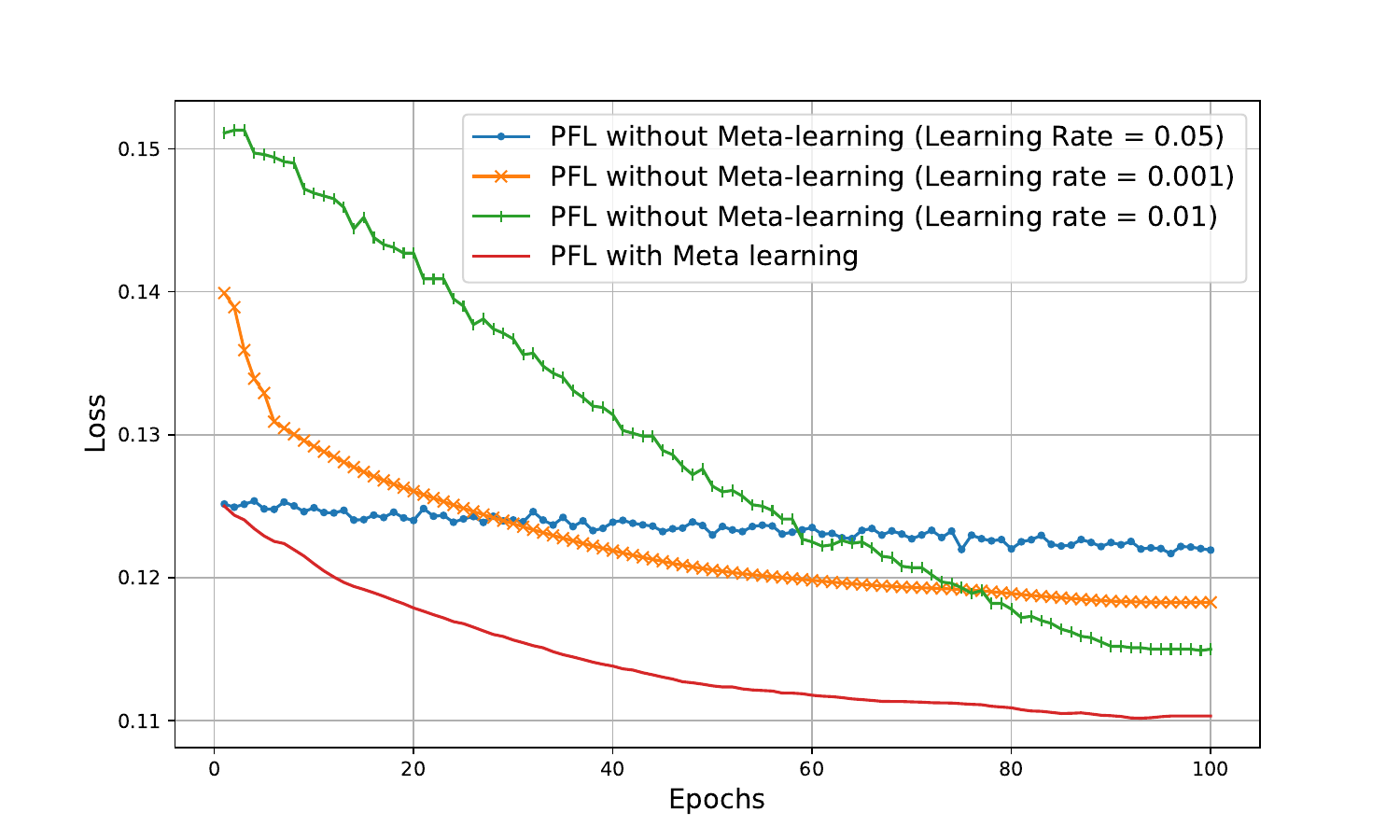} &
      \includegraphics[width=.99\columnwidth]{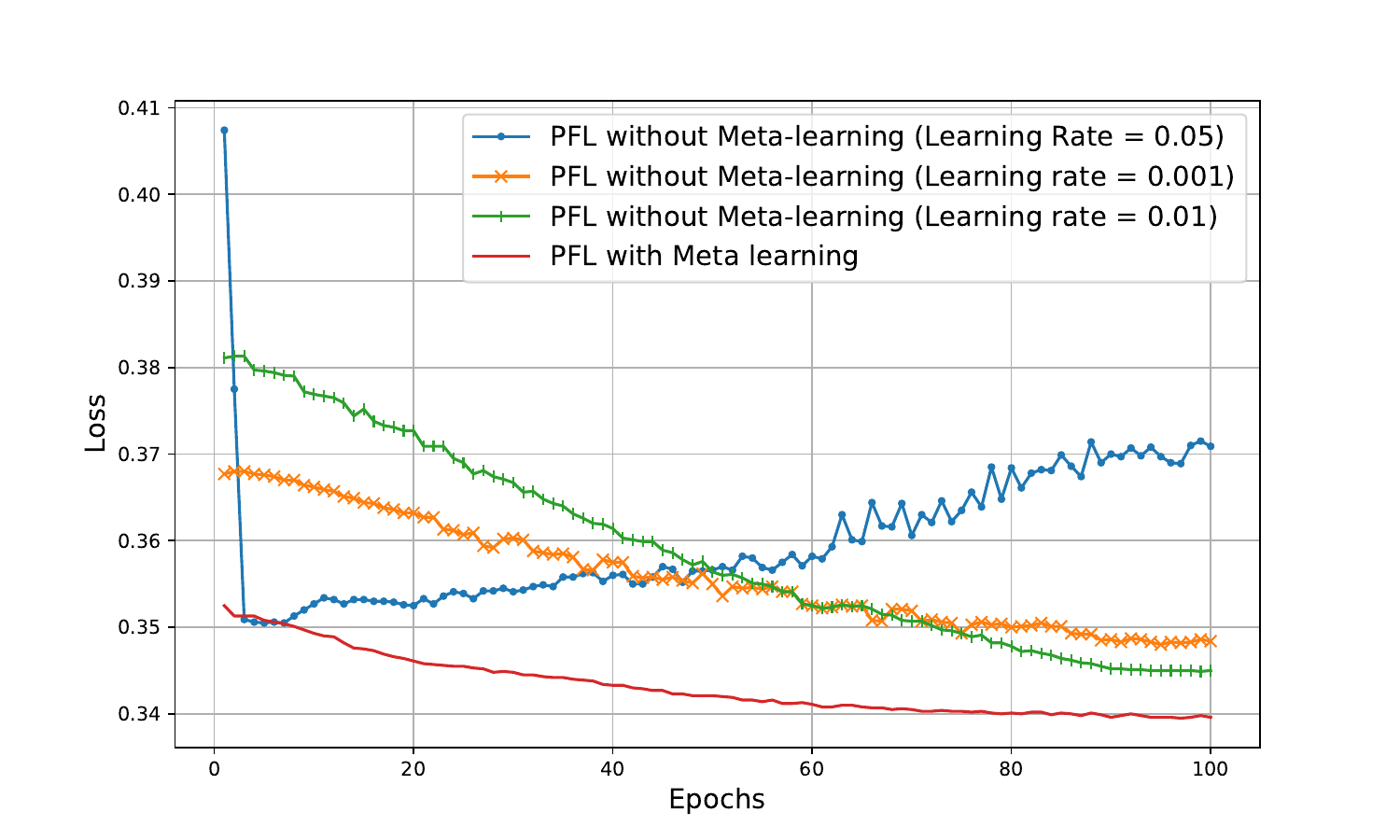} \\
    \small (a) MAE Loss &
      \small (b) RMSE Loss\\
  \end{tabular}
  \medskip
  \vspace{-5mm}
  \caption{Comparison between different learning rates and meta-learning for 5 clients non-IID data.}
  \label{fig: lr}
  \vspace{-5mm}
\end{figure*}
The proposed PFL approach is summarized in Algorithm~\ref{algo: metaSGD}. In this algorithm, in every global round, after SM obtains the global weight in line 7, they perform meta-learning functionalities in lines 8-12. Here, the loss value is calculated for every available learning rate ($\alpha_j$) (line 9). The $\alpha_j$ that produces the lowest loss value is then returned to the local model as the optimal learning rate in line 11. Then in line 13, we perform the local model training and send the updated local model to the server (line 15) after $T$ rounds of local rounds. Then the server does federated averaging on line 17, testing on line 18, and then saves the updated weight for the next global round. Finally after $K$ global rounds, we get our optimal global model $(\boldsymbol{w}^*)$. This approach allows us to use different learning rates that enable SMs with good datasets and better performance to use a different learning rate than those with poor performance and datasets. As a result, the global model is impacted separately for every SM and the global model can have more accurate updates by using the personalized factor.

\section{Simulations and Performance Evaluation}
\subsection{Environment Settings}
For our load forecasting simulations, we used the \textit{Individual Household Electric Power Consumption} dataset \cite{hebrail2012individual}. It is both multivariate and time-series real-life data that is focused on physics and chemistry that describes the electricity consumption for a single household over 47 months, from December 2006 to November 2010. The house is located in Sceaux, 7km away from Paris, France. It has a total of 2075259 instances with 9 features. From the features, there are a total of seven variables and two other non-variables: Date and Time. The variables include: 
\begin{itemize}
    \item global\_active\_power: Household consumption of total active power consumed
    \item global\_reactive\_power: Household consumption of total reactive power consumed
    \item voltage: The average voltage (volts) in that household 
    \item global\_intensity: The average intensity (apms) in that household
    \item sub\_metering\_1: Kitchen's active energy (watt-hours)
    \item sub\_metering\_2: Laundry's active energy (watt-hours)
    \item sub\_metering\_3: Climate control system's active energy (watt-hours)
\end{itemize} 
There are multiple layers in the design of the LSTM model for load prediction. Sequences of shape (batch\_size, 24, 1) are first entered into the input layer; each sequence comprises 24 hourly load values. An LSTM layer comprising 50 units comes next, processing the sequential input and capturing temporal dependencies. The next step is to add a dropout layer, whose dropout rate is 0.2. During training, it randomly sets 20\% of the LSTM layer's outputs to zero in order to prevent over-fitting. If more complicated patterns need to be captured, an additional LSTM layer can be added after this. Lastly, the expected load value for the following hour is generated by adding a completely connected dense layer with a single neuron. This architecture uses the long-term relationships that the LSTM can maintain in the data, making it appropriate for precise load forecasting.

\subsection{Client Distribution}
For our research, first, we compared our results with 2, 3, and 5 clients IID data distribution. Then we implemented the comparison using 5 clients in the non-IID data distribution. Here, the clients have different batch sizes and number of data. For our simulations, the batch sizes are 128,128,128,64, and 256, and for the data points five clients have 20\%, 20\%, 20\%, 10\%, and 25\% data. For example, client 2 has a batch size of 128 and has 415051 data instances. So there is data heterogeneity and unequal data amounts for non-iid data distribution. For training the model and creating a gradient, we considered three different learning rates 0.05, 0.001, and 0.0001. For every global epoch, a meter had the ability to temporarily test its performance based on all three of these learning rates for 10 local rounds and then judging by the performance we selected the optimal learning rate (the learning rate that produces the lowest loss value) and use that for local training for 50 local rounds. we run our simulation results for 100 global rounds. 

\subsection{Simulation Results}
\subsubsection{Training Performance}
At first, we implement our proposed personalized Meta-LSTM FL algorithm in IID settings for different clients. Fig. \ref{fig: iid} describes the result for both MSE loss (a) and RMSE loss (b) for 2,3, and 5 clients as well as standalone and centralized schemes. 

The figure shows that, as expected, the centralized scheme performs the best while the standalone performs the worst for both loss values. Additionally, the loss value reduces with an increase in clients, suggesting improved performance. As a result, we draw the conclusion that our approach is adaptable for more clients and is proportionate to the number of clients. The following simulation results have been obtained using non-IID data from five clients. Fig. \ref{fig: lr} shows the simulation result for the usage of different learning rates and meta-learning.  The reason why meta-learning works better than other individual learning methods is evident in that figure. While learning rates at 0.05 get off to the best starts but fail to maintain and overestimate both loss values. Learning rates at 0.001 is the too slow pace to catch up and communication overhead grows. And learning rate 0.01 did not have a good start to catch up. Nevertheless, meta-learning is able to extract all the beneficial features from these learning rates through appropriate use, which makes it perfect for more accurate results with less communication overhead. 

\subsubsection{Load Forecasting Performance}
\begin{figure*}
  \centering
  \begin{tabular}{ c @{\hspace{3pt}} c }
    \includegraphics[width=.99\columnwidth]{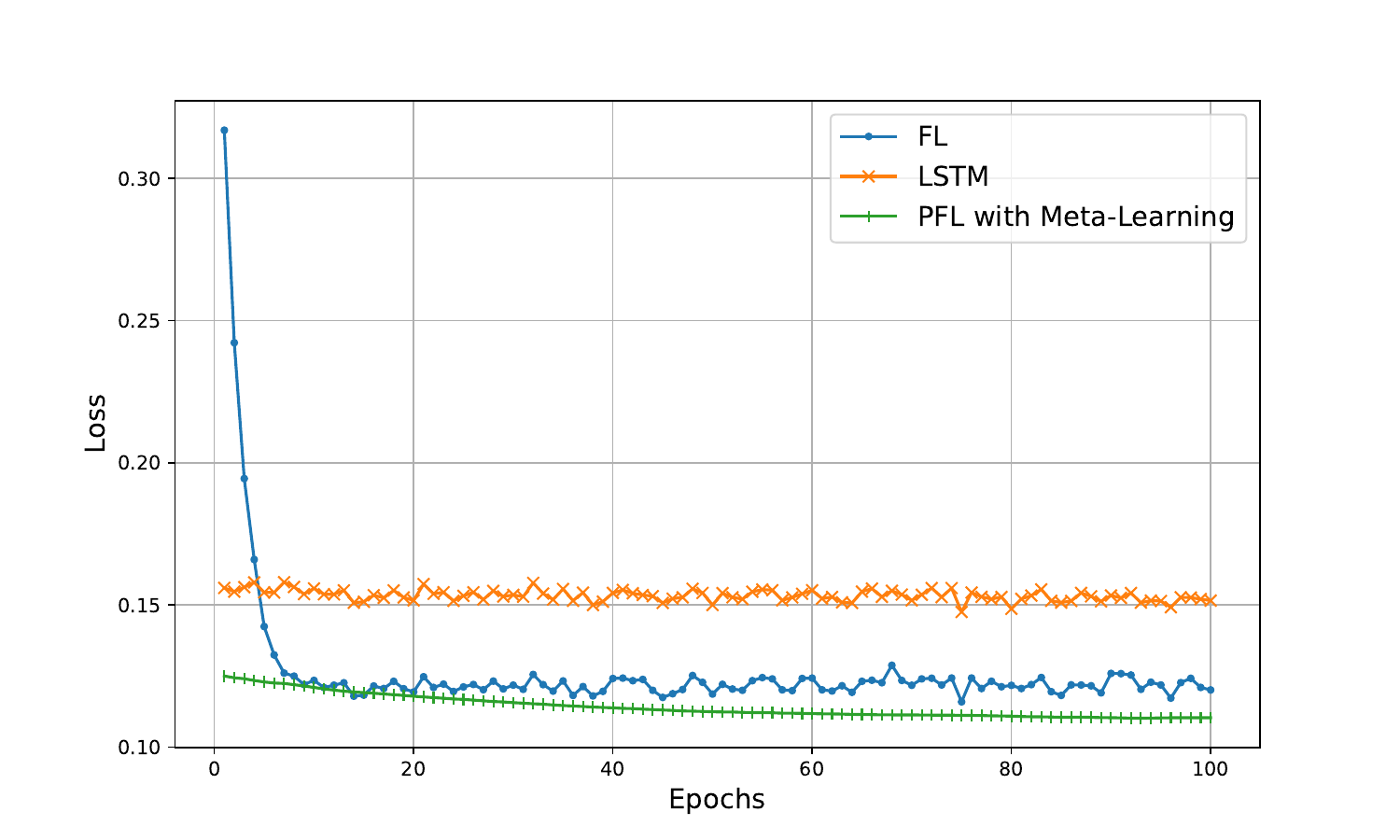} &
      \includegraphics[width=.99\columnwidth]{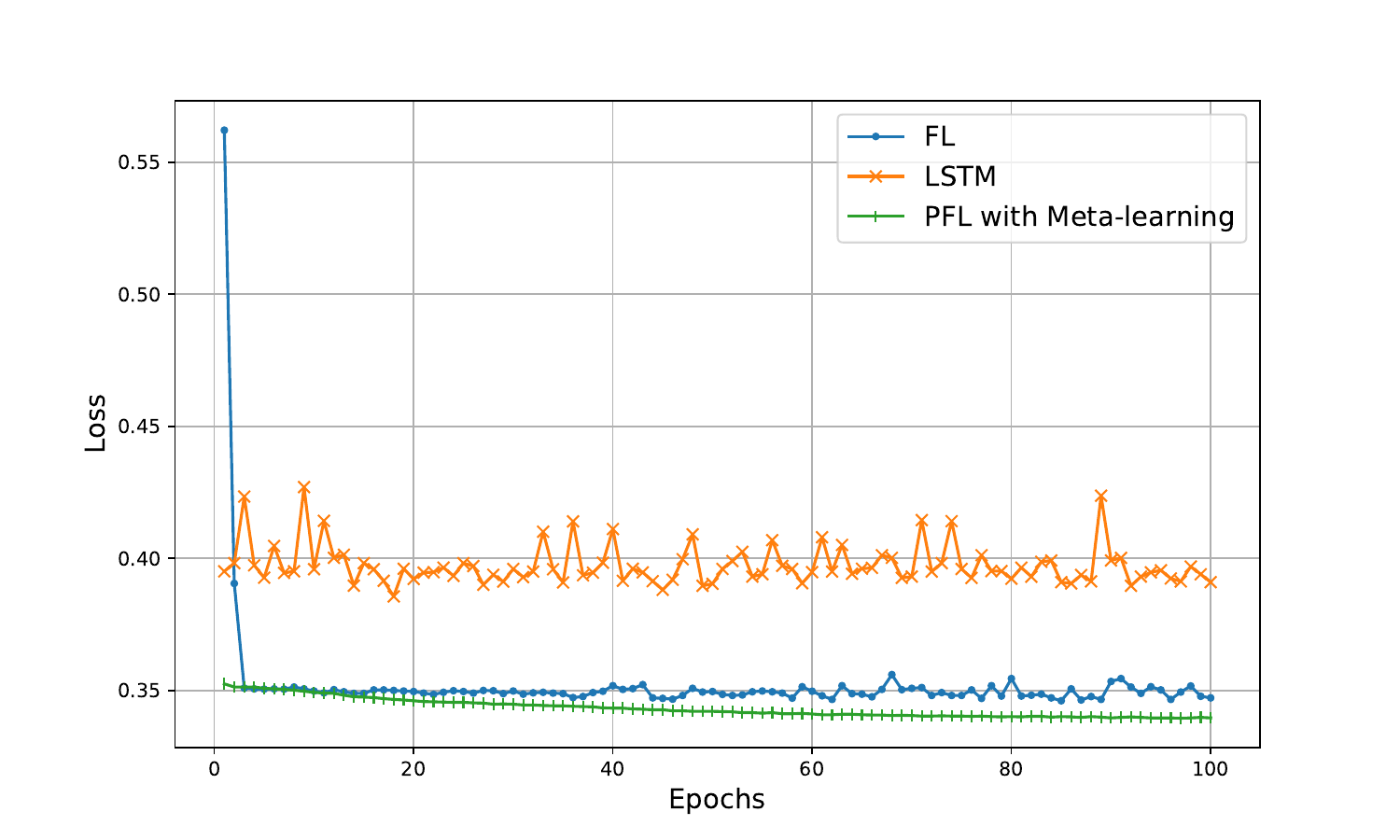} \\
    \small (a) MAE Loss &
      \small (b) RMSE Loss\\
  \end{tabular}
  \medskip
  \vspace{-5mm}
  \caption{Comparison between state-of-the-art approaches (LSTM and FL) and our approach.}
  \label{fig: comp}
  \vspace{-1mm}
\end{figure*}

Finally, in Fig \ref{fig: comp} we have compared our performance with the LSTM algorithm \cite{bouktif2018optimal} and state of art FL algorithm \cite{fekri2022distributed} based on global epochs. The graph shows that, among the three methods, LSTM performed the worst, as would be expected. PFL fared somewhat better than FL in this comparison. Furthermore, yields a far more stable result than FL. 
\begin{figure*}
  \centering
  \begin{tabular}{ c @{\hspace{2pt}} c @{\hspace{2pt}} c}
    \includegraphics[width=.66\columnwidth]{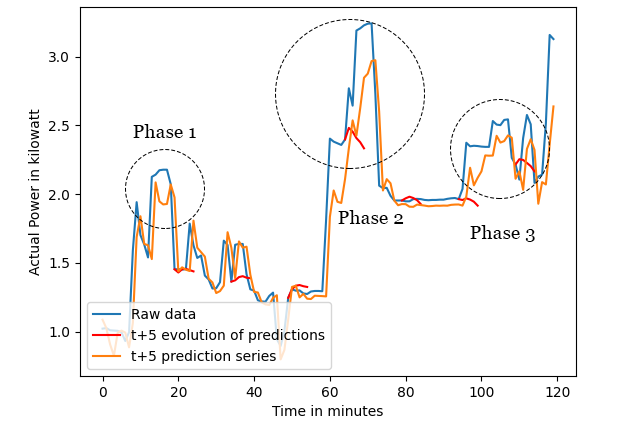} &
    \includegraphics[width=.66\columnwidth]{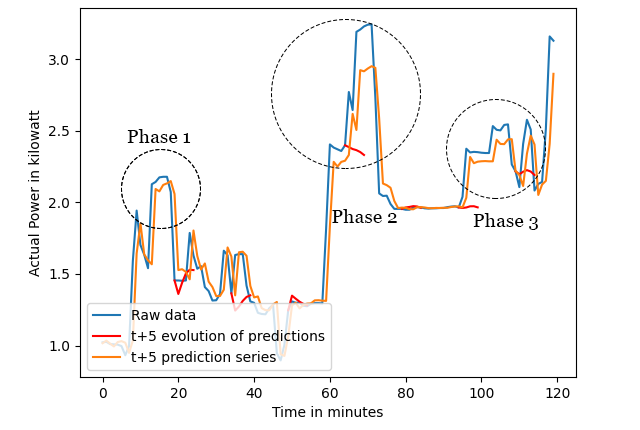} &
      \includegraphics[width=.66\columnwidth]{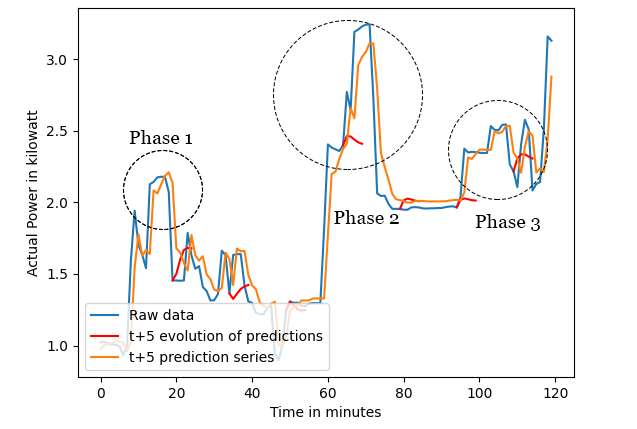} \\
    \small (a) LSTM Result  &
      \small (b) FL Result &
      \small (c) PFL Result
  \end{tabular}
  \medskip
  \vspace{-5mm}
  \caption{Simulation result of the original value and predicted value for the first 120 minutes in the testing dataset.}
  \label{fig: ovp}
  \vspace{-5mm}
\end{figure*}
We present the testing findings for the first 120 minutes for each of the three ways in Fig. \ref{fig: ovp} to conduct additional investigation. More precisely, we concentrate on the three crucial stages denoted as phases 1, 2, and 3. In this case, the orange line indicates the expected outcome while the blue line represents the raw data. As a result, the loss value is represented by the space and variations between the orange and blue lines. From Fig \ref{fig: ovp}(a) we can observe that the LSTM technique is not very good at predicting changes, particularly in phases 2 and 3, which have a large number of curves. In phase 1, FL and PFL are far more accurate at predicting. Still, they have not been able to accurately forecast phase 2's peak. Furthermore, we can observe the variations in phase 2 and particularly in phase 3 if we compare the FL and PFL results in Figs. \ref{fig: ovp}(b-c). PFL was able to achieve both the highest point in phase 2 and noticeably improved performance in phase 3. 

\section{Conclusion}
This paper has proposed a novel PFL approach called the personalized meta-LSTM algorithm for load prediction for non-IID data. Our model has a flexible meter participation technique allowing clients with various configurations to join in the learning process. The proposed meta-learning solution within our PFL framework enables removing any additional server costs, allowing communication to be more efficient and reliable. Our simulation result also proves that meta-learning-based PFL performed better than both LSTM an FL approaches. By utilizing more advanced FL techniques, we can further improve the performance by either improving the accuracy or making data more secure. In future work, we will consider incentive mechanisms to further speed up load forecasting at SMs.  

\ifCLASSOPTIONcaptionsoff
  \newpage
\fi
\bibliographystyle{ieeetr}
\bibliography{bibtex/bib/IEEEexample}
\end{document}